\documentclass{article} 
\usepackage{iclr2021_conference,times}

\usepackage{hyperref}
\usepackage{url}

\usepackage{smile}

\usepackage{graphicx}

\title{Beyond Fully-Connected Layers with \\Quaternions: Parameterization of Hypercomplex Multiplications with $1/n$ Parameters}

\author{Aston Zhang$^\dagger$, Yi Tay$^\ddagger$\thanks{Work was done at NTU.}, Shuai Zhang$^\diamond$, Alvin Chan$^\triangleleft$ \\\textbf{Anh Tuan Luu$^{\triangleright}$, Siu Cheung Hui$^\triangleleft$, Jie Fu}$^\bullet$\\
$^\dagger$Amazon Web Services AI\\ $^\ddagger$Google Research\\ $^\diamond$ETH Zürich\\ $^\triangleleft$NTU, Singapore\\ $^\triangleright$VinAI Research\\ $^\bullet$Mila, Université de Montréal\\
\texttt{az@astonzhang.com} \\
}

\iclrfinalcopy 
\begin{document}

\maketitle

\begin{abstract}
Recent works have demonstrated reasonable success of representation learning in hypercomplex space. 
Specifically, 
``fully-connected layers with Quaternions'' (4D hypercomplex numbers),
which replace real-valued matrix multiplications
in fully-connected layers
with Hamilton products of Quaternions,
both enjoy parameter savings
with only $1/4$ learnable parameters
and achieve comparable performance in various applications.
However,
one key caveat is that hypercomplex space 
only exists at very few predefined dimensions (4D, 8D, and 16D).
This
restricts the flexibility of models that leverage hypercomplex multiplications.
To this end, we propose parameterizing hypercomplex multiplications,
allowing models to learn multiplication rules
from data regardless of whether such rules are predefined.
As a result,
our method not only subsumes the Hamilton product,
but also learns to operate
on any arbitrary $n$D hypercomplex space, 
providing more architectural flexibility
using arbitrarily $1/n$ learnable parameters compared with the fully-connected layer counterpart.
Experiments 
of applications to the LSTM and Transformer models
on natural language inference, machine translation, 
text style transfer, and subject verb agreement
demonstrate architectural flexibility and effectiveness of the proposed approach.
\end{abstract}
\section{Introduction}
\label{sec_intro}

A Quaternion is a 4D hypercomplex number with one real component and three imaginary components.
The Hamilton product is the hypercomplex multiplication
of two Quaternions.
Recent works in Quaternion space and Hamilton products have demonstrated reasonable success~\citep{parcollet2018quaternion,parcollet2019quaternion,tay2019lightweight}.
Notably, the Hamilton product
enjoys 
a parameter saving with $1/4$ learnable parameters
 as compared with the real-valued matrix multiplication.
 It also enables effective representation learning by modeling interactions between real and imaginary components.

One of the attractive properties of Quaternion models is its high applicability and universal usefulness to one of the most ubiquitous layers in deep learning, i.e., the  fully-connected (or feed-forward) layer.
Specifically, 
``fully-connected layers with Quaternions'' replace real-valued matrix multiplications
in fully-connected layers
with Hamilton products of Quaternions,
enjoying parameter savings
with only $1/4$ learnable parameters
and achieving comparable performance with their fully-connected layer counterparts~\citep{parcollet2018quaternion,parcollet2019quaternion,tay2019lightweight}.

The fully-connected layer is one of the most dominant components in existing deep learning literature~\citep{Goodfellow-et-al-2016,zhang2020dive}.
Its pervasiveness cannot be understated, given its centrality to many core building blocks in neural network research.
Given widespread adoptions of fully-connected layers, e.g., within LSTM networks ~\citep{hochreiter1997long} and Transformer models ~\citep{vaswani2017attention}, 
having flexibility to balance
between parameter savings and effectiveness
could be extremely useful to many real-world applications.

Unfortunately, hypercomplex space only exists at 4D (Quaternions), 8D (Octonions),
and 16D (Sedonions), which generalizes the 2D complex space~\citep{rishiyur2006neural}.
Moreover, custom operators are required at each hypercomplex dimensionality.
For instance, 
the Hamilton product is the hypercomplex multiplication
in 4D hypercomplex space.
Thus, no operator in such predefined hypercomplex space
is suitable
for applications that prefer reducing parameters to $1/n$,
where $n \neq 4, 8, 16$.

In view of the architectural limitation 
due to the very few choices of those existing hypercomplex space,
we propose parameterization of hypercomplex multiplications,
i.e., learning the real and imaginary component interactions from data in a differentiable fashion. Essentially, our method can operate on an arbitrary $n$D hypercomplex space,
aside from 
subsuming those predefined hypercomplex multiplication rules,
facilitating 
using up to \emph{arbitrarily} $1/n$ learnable parameters
while maintaining expressiveness. 
In practice, the hyperparameter $n$
can be flexibly specified or tuned
by users based on applications.



Concretely, our prime contribution is a new module that parameterizes and generalizes the hypercomplex multiplication by learning the real and imaginary component interactions,
i.e., multiplication rules, from data. Our method, which we call the parameterized hypercomplex multiplication layer, is characterized by a sum of Kronecker products that generalize the vector outer products to higher dimensions in real space. To demonstrate applicability, we equip two well-established models (the LSTM and Transformer) with our proposed method. We conduct extensive experiments on different tasks, i.e., natural language inference for LSTM networks and machine translation for Transformer models. Additionally, we perform further experiments on text style transfer and subject verb agreement tasks.
All in all, our method 
has demonstrated architectural flexibility
through different experimental settings,
where it generally 
can use a fraction of the learnable parameters
with minimal degradation or slight improvement in performance.



The overall contributions of this work are summarized as follows:
\begin{itemize}
    \item We propose a new parameterization of hypercomplex multiplications: the parameterized hypercomplex multiplication (PHM) layer. 
    This layer has $1/n$ learnable parameters compared with the fully-connected layer counterpart, where $n$ can be flexibly specified by users.
    The key idea behind PHM layers is to learn the 
    interactions between real and imaginary components, i.e., multiplication rules, from data using a sum of Kronecker products.
    \item We demonstrate the applicability of the PHM layers by leveraging them in two dominant neural architectures: the LSTM and Transformer models.
    \item We empirically show architectural flexibility and effectiveness of PHM layers by conducting extensive experiments on five natural language inference tasks, seven machine translation datasets, together with text style transfer and subject verb agreement tasks. 
\end{itemize}

\section{Background on Quaternions and Hamilton Products}
We begin by introducing the background for the rest of the paper. Concretely, we describe Quaternion algebra along with Hamilton products, which is at the heart of our proposed approach.
\paragraph{Quaternion} A Quaternion $Q \in \mathbb{H}$ is a hypercomplex number with one real component and three imaginary components as follows:
\begin{align}
\label{eq:qq}
Q = Q_r + Q_x\mathbf{i} + Q_y\jb + Q_z\kb,
\end{align}
whereby $\textbf{ijk}= \mathbf{i}^2=\jb^2=\kb^2=-1$. In \eqref{eq:qq}, noncommutative multiplication rules hold: $\mathbf{ij} = \kb, \mathbf{jk} = \mathbf{i}, \mathbf{ki} = \jb, \mathbf{ji} = -\kb, \mathbf{kj} = -\mathbf{i}, \mathbf{ik} = -\jb$.  Here, $Q_r$ is the real component, $Q_x,Q_y,Q_z$ are real numbers that represent the imaginary components of the Quaternion $Q$. 

\paragraph{Addition} The addition of two Quaternions is defined as
\begin{align*} 
Q+P=
Q_r + P_r + (Q_x + P_x)\mathbf{i} + (Q_y + P_y)\jb + (Q_z + P_z)\kb,
\end{align*}
where $Q$ and $P$ with subscripts denote the real and imaginary components of Quaternions $Q$ and $P$.
\paragraph{Scalar Multiplication} Any scalar $\alpha$ multiplies across all the components:
\begin{align*}
\alpha Q = \alpha Q_r + \alpha Q_x \mathbf{i} + \alpha Q_y \jb +\alpha Q_z \kb.
\end{align*}
\paragraph{Hamilton Product} The Hamilton product, which represents the multiplication of two Quaternions $Q$ and $P$, is defined as
\begin{align}
\label{eq:hamilton}
Q \otimes P &= (Q_rP_r - Q_xP_x - Q_yP_y - Q_zP_z) + (Q_xP_r + Q_rP_x - Q_zP_y + Q_yP_z)\:\mathbf{i} \nonumber
         \\ &+ (Q_yP_r + Q_zP_x + Q_rP_y - Q_xP_z)\:\jb + (Q_zP_r - Q_yP_x + Q_xP_y + Q_rP_z)\:\kb.
\end{align}
The multiplication rule in \eqref{eq:hamilton}
forges interactions between real and imaginary components of $Q$ and $P$. The benefits of Hamilton products have been demonstrated in recent works where the matrix multiplication in fully-connected layers is replaced with the Hamilton product: 
this reduces 75\%  parameters with comparable performance~\citep{parcollet2018quaternion,parcollet2019quaternion,tay2019lightweight}.
 
\section{Parameterization of Hypercomplex Multiplications}
The following introduces our proposed parameterized hypercomplex multiplication  layer and elaborates on how it parameterizes and generalizes 
multiplications in hypercomplex space,
such as subsuming the multiplication rules of Hamilton products in \eqref{eq:hamilton}.

\subsection{Fully-Connected (FC) Layers}
Before we delve into our proposed method, recall the fully-connected (FC) layer that transforms an input $\xb \in \RR^d$ into an output $\yb \in \RR^k$ by
\begin{align}
\label{eq:fc}
\yb = \text{FC}(\xb) = \Wb \xb + \bb,
\end{align}
where the weight matrix of parameters $\Wb \in \RR^{k \times d}$ and the bias vector of parameters $\bb \in \RR^{k}$.
The FC layer in \eqref{eq:fc} is fundamental to many modern and traditional neural network architectures. 
Note that the degree of freedom for the weight parameters $\Wb$ in \eqref{eq:fc} is $kd$. Since $\Wb$ dominates parameterization, the parameter size of the FC layer in \eqref{eq:fc} is $\cO(kd)$.

\subsection{Parameterized Hypercomplex Multiplication (PHM) Layers} 
\label{subsec:phm layers}

We propose the parameterized hypercomplex multiplication (PHM) layer
that transforms an input $\xb$ into an output $\yb$ by
\begin{align}
\label{eq:PHM}
\yb = \text{PHM}(\xb) = \Hb \xb + \bb,
\end{align}
where 
the same notation from \eqref{eq:fc} is used but 
the replaced parameter $\Hb \in \RR^{k \times d}$
is constructed 
by a sum of Kronecker products.
For context,
the Kronecker product is a generalization of the vector outer product to higher dimensions in real space.
For any matrix $\Xb \in \RR^{m \times n}$ and $\Yb \in \RR^{p \times q}$,
the Kronecker product $\Xb \otimes \Yb$ is a block matrix:

\begin{align*}
\Xb \otimes \Yb = \begin{bmatrix} 
    x_{11}\Yb & \dots  & x_{1n}\Yb \\
    \vdots & \ddots & \vdots \\
    x_{m1}\Yb & \dots  & x_{mn}\Yb 
    \end{bmatrix} \in \RR^{mp \times nq},
\end{align*}
where $x_{ij}$ is the element of $\Xb$ at its $i^{\text{th}}$ row and $j^{\text{th}}$ column. Note that the symbol $\otimes$ between two matrices is the Kronecker product while the same symbol between two Quaternions  means the Hamilton product.

\begin{figure*}[t]
    \centering
  \includegraphics[width=0.95\linewidth]{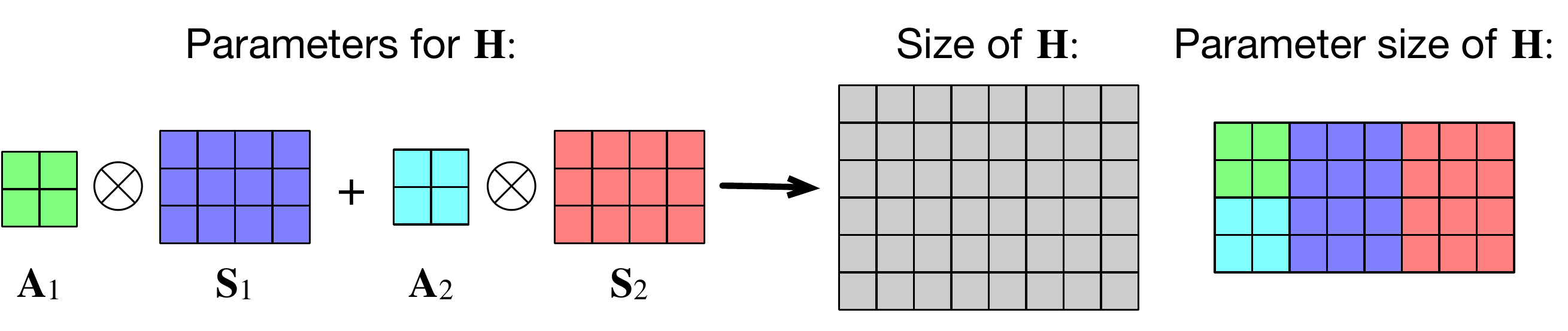}
  \vspace{-1em}
    \caption{Illustration of the PHM layer. It uses a sum of Kronecker products of matrices $\Ab_i$ and $\Sbb_i$ ($i=1,2$) to construct $\Hb$ in \eqref{eq:PHM} (here $n=2, k=6, d=8$). Best viewed in color.}
    \label{fig:kronecker}
\end{figure*}

Now let us revisit \eqref{eq:PHM} to explain $\Hb$.
Suppose that both $k$ and $d$ are divisible by a user-defined hyperparameter
$n \in \ZZ_{> 0}$.
For $i = 1, \ldots, n$, denote by each parameter matrix
$\Ab_i \in \RR^{n \times n}$ and $\Sbb_i \in \RR^{\frac{k}{n} \times \frac{d}{n}}$. 
The parameter $\Hb$ in \eqref{eq:PHM} is 
a sum of $n$ Kronecker products:
\begin{align}
\label{eq:H_kron}
\Hb = \sum_{i=1}^n \Ab_i \otimes \Sbb_i.
\end{align}
As illustrated in Figure~\ref{fig:kronecker},
it is the parameter matrices $\Ab_i$ and $\Sbb_i$ ($i = 1, \ldots, n$) that determine the degree of freedom for $\Hb$, which is $kd/n + n^3$.
Since $\Hb$ dominates parameterization, the parameter size of the PHM in \eqref{eq:PHM} is $\cO(kd/n)$, where $kd \gtrapprox n^4$ is assumed: this condition is mild for real-world problems, such as in our experiments (e.g., $d = 512$, $k=2048$, $n=2,4,8,16$). Thus, for the same input and output sizes,
the parameter size of a PHM layer is approximately $1/n$ of that of an FC layer under mild assumptions.

The benefit of parameterization reduction of PHM layers is
due to reusing elements of both parameter matrices $\Ab_i$ and $\Sbb_i$
in the Kronecker product.
As an alternative perspective,
we can equivalently reconstruct $\Hb$ in \eqref{eq:H_kron}
by reusing parameter matrices in real-valued matrix multiplications,
followed by more operations.
Due to limited space,
this more complicated perspective is offered in Appendix~\ref{sec:Reconstructing the Parameter Matrix}.
Though simply setting $\Hb = \Ab_1 \otimes \Sbb_1$
can further save parameters,
it does not generalize hypercomplex multiplications
hence is out of scope.

\begin{figure}[t]
\centering
\subfigure[Learning rotations in 3D real space]{
\includegraphics[scale=0.4]{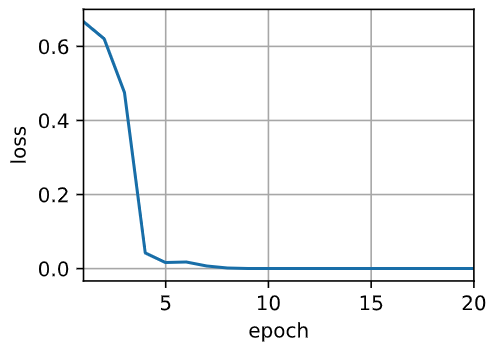}
\label{fig:rotation}
}
\subfigure[Learning Hamilton products in Quaternion space]{
\includegraphics[scale=0.4]{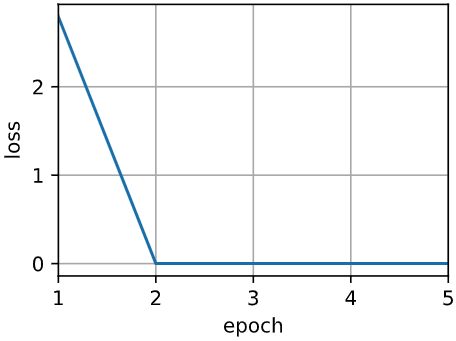}
\label{fig:phm-hamilton}
}
\label{fig:learning-phm}
\caption{PHM layers can learn to perform rotations in 3D real space and Hamilton products in Quaternion space on artificial datasets.}
\end{figure}

To show that PHM layers 
can learn to perform pre-defined multiplication-related operations in practice,
we perform experiments 
to learn rotations in 3D real space using the PHM layer.
Using a rotation matrix $\Wb \in \RR^{3 \times 3}$
we create an artificial dataset
$\{(\xb_i \in \RR^{3}, \yb_i \in \RR^{3})\}$,
where $\yb_i$ is generated via the 3D rotation of the input:
$\yb_i = \Wb \xb_i$. Figure~\ref{fig:rotation} shows that the loss converges to zero:
the PHM layer can learn a single rotation of an object in 3D real space.

In the following,
we show how the proposed PHM layer
subsumes and generalizes
both hypercomplex multiplications and real-valued matrix multiplications.

\subsection{Subsuming Hypercomplex Multiplications}
\label{sec:hyper}

First, we explore how the PHM layer connects to the hypercomplex multiplication. For the sake of illustration, 
let us take the Hamilton product of two Quaternions $Q$ and $P$ in \eqref{eq:hamilton}
as an example, which can be rewritten as
\begin{align}
\label{eq:hamilton_kron}
\begin{bmatrix}
    Q_r & -Q_x & -Q_y & -Q_z \\
    Q_x & Q_r & -Q_z & Q_y \\
    Q_y & Q_z & Q_r & -Q_x \\
    Q_z & -Q_y & Q_x & Q_r \\
\end{bmatrix} 
\begin{bmatrix}
    P_r \\
    P_x\\
    P_y\\
    P_z \\ 
\end{bmatrix},
\end{align}
where the 4 output elements are the real values for the Quaternion unit basis $[1, \ib, \jb, \kb]^\top$. 
Note that for models
leveraging Hamilton products of Quaternions
\citep{parcollet2018quaternion,parcollet2019quaternion,tay2019lightweight},
the components $Q_r, Q_x, Q_y, Q_z$ of
\eqref{eq:hamilton_kron} are learnable parameters
while the components $P_r, P_x, P_y, P_z$ are the layer inputs.
In practice, such a layer usually has more than 4 inputs ($d > 4$).
To apply the Hamilton product,
all the inputs are evenly split
into 4 segments ($P_r, P_x, P_y, P_z$) of the right input vector of \eqref{eq:hamilton_kron}.
Then each component in the left matrix of \eqref{eq:hamilton_kron}
can be a block matrix (i) where all the elements take the same value;
(ii) whose shape is aligned with the input length $d$ and 
the output length $k$ of the layer.
It is noteworthy that the left $4\times 4$ matrix of 
\eqref{eq:hamilton_kron} can be rewritten as a sum of 4 Kronecker products:
\begingroup\makeatletter\def\f@size{7}
\begin{align}
\label{eq:ASQ_kron}
\underbrace{
\begin{bmatrix}
    1 & 0 & 0 & 0 \\
    0 & 1 & 0 & 0 \\
    0 & 0 & 1 & 0 \\
    0 & 0 & 0 & 1 \\
\end{bmatrix}
}_{\Ab_1}
\otimes 
\underbrace{
\begin{bmatrix}
    Q_r \\
\end{bmatrix}
}_{\Sbb_1}
+
\underbrace{
    \begin{bmatrix}
    0 & -1 & 0 & 0 \\
    1 & 0 & 0 & 0 \\
    0 & 0 & 0 & -1 \\
    0 & 0 & 1 & 0 \\
\end{bmatrix}
}_{\Ab_2}
\otimes 
\underbrace{
\begin{bmatrix}
    Q_x \\
\end{bmatrix}
}_{\Sbb_2}
+
\underbrace{
    \begin{bmatrix}
    0 & 0 & -1 & 0 \\
    0 & 0 & 0 & 1 \\
    1 & 0 & 0 & 0 \\
    0 & -1 & 0 & 0 \\
\end{bmatrix}
}_{\Ab_3}
\otimes 
\underbrace{
\begin{bmatrix}
    Q_y \\
\end{bmatrix}
}_{\Sbb_3}
+
\underbrace{
    \begin{bmatrix}
    0 & 0 & 0 & -1 \\
    0 & 0 & -1 & 0 \\
    0 & 1 & 0 & 0 \\
    1 & 0 & 0 & 0 \\
\end{bmatrix}
}_{\Ab_4}
\otimes
\underbrace{
\begin{bmatrix}
    Q_z \\
\end{bmatrix}
}_{\Sbb_4}
.
\end{align}
\endgroup

According to \eqref{eq:ASQ_kron}, when $n=4$,
the PHM layer can be learned to express the Hamilton product of Quaternions.
Specifically, matrices $\Ab_1, \ldots, \Ab_4$ in \eqref{eq:H_kron} parameterize
the four matrices composed of $-1, 0, 1$ in \eqref{eq:ASQ_kron} that reflect interactions
between real and imaginary components of Quaternions,
which are the rule of Hamilton products.
The single-element ``matrices'' $\Sbb_1, \ldots, \Sbb_4$ in \eqref{eq:H_kron}
are equal to the learnable components $Q_r, Q_x, Q_y, Q_z$ in \eqref{eq:hamilton_kron}.
Figure~\ref{fig:phm-hamilton} shows that PHM layers can learn the rule of Hamilton products on artificial data.
Likewise, hypercomplex multiplications of Octonions or Sedenions can also be learned by the PHM layer when $n$ is set to 8 or 16.

\subsection{Subsuming Real-Valued Matrix Multiplications}

Next, we show how the PHM layer subsumes the matrix multiplication in real space.
In other words, the PHM layer is a generalization of the FC layer via the hyperparameter $n$.
To explain, referring to \eqref{eq:PHM}, when $n=1$, $\Hb = \Ab_1 \otimes \Sbb_1 = a \Sbb_1$, 
where the scalar $a$ is the single element of the $1\times 1$ matrix $\Ab_1$ and 
$\Sbb_1 \in \RR^{k \times d}$.
Since learning $a$ and $\Sbb_1$ separately is equivalent to learning their multiplication jointly, scalar $a$ can be dropped,
which is learning the single weight matrix in an FC layer. Therefore, a PHM layer is degenerated to an FC layer when $n=1$.

\subsection{Generalizing Hypercomplex Multiplications}

Though parameter reusing by component-wise partitioning in Quaternion space has demonstrated  success~\citep{parcollet2018quaternion,zhu2018quaternion,parcollet2019quaternion,tay2019lightweight},
one key problem is that 
hypercomplex space only exists at very few predefined  dimensionalities, 
such as 4D (Quaternions), 8D (Octonions), and 16D (Sedonions).
Within the context of hypercomplex space,
specialized multiplication rules, such as the Hamilton product,
have to be devised and encoded in the network as a fixed inductive bias. 
As described in 
Section~\ref{sec_intro},
the very few choices over existing hypercomplex space
restricts the flexibility of networks that
leverage hypercomplex multiplication.

In sharp contrast to relying on predefined mathematical rules
over limited dimensionality choices, 
the PHM layer 
treats the dimensionality
$n$ (number of Kronecker products) as a tunable hyperparameter
and learns such specialized multiplication rules from data, as manifested in the parameterized matrices $\Ab_i$ ($i = 1, \ldots, n$) in \eqref{eq:H_kron}.
On one hand, the PHM layer can express hypercomplex multiplications when $\Ab_i$ are set to reflect those predefined multiplication rules in hypercomplex space.
On the other hand, the  PHM layer can be seen as a trainable and parameterized form of $n$D hypercomplex multiplications, where $n$ can be values other than 4, 8, or 16.
Thus, the PHM layer generalizes  multiplications in hypercomplex space.
Since $n$ can be 1,
the PHM layer also offers a neat way to bridging multiplication between both real space and hypercomplex space.

\section{Neural Models with PHM Layers}
To demonstrate the applicability of the PHM layers,
we develop the PHM-LSTM and PHM-Transformer by
equipping two popular neural network models, LSTMs and Transformers, with PHM layers.

\subsection{PHM-LSTM}

Recurrent neural networks such as LSTMs~\citep{hochreiter1997long} are gated
recurrent networks where the gating functions are parameterized by linear transformations. We introduce the PHM-LSTM, which replaces such linear transformations in LSTMs with PHM layers:
\begin{align*}
\yb_t &= \text{PHM }(\xb_t) + \text{PHM }(\hb_{t-1}) + \bb\\
\fb_t, \ib_t, \ob_t, \xb'_t &= \phi(\yb_t) \\
\cbb_t &= \sigma_s(\fb_t) \: \cbb_{t-1} + \sigma_s(\ib_t) \: \sigma_{t}(\xb'_t) \\
\hb_t &= \ob_t \odot \cbb_t,
\end{align*}
where $\sigma_s$ is the sigmoid activation function, $\sigma_{t}$ is the tanh activation function, $\phi: \RR^{1  \times d} \rightarrow \RR^{4 \times \frac{d}{4}}$ is a four-way split on the last dimension, and $\cbb_{t}, \hb_{t}$ are the cell state and the hidden state of the PHM-LSTM unit at any time step $t$. 

\subsection{PHM-Transformer}

The Transformer is a stacked neural network architecture that aggressively exploits linear transformations~\citep{vaswani2017attention}. Each self-attention layer comprises of $\Qb$ (query), $\Kb$ (key), $\Vb$ (value) linear transformations, along with multiple heads. Each Transformer block also has a position-wise feed-forward network composed of two FC layers.
Since a large majority of the Transformer parameters stem from linear transformations or FC layers,
we introduce the PHM-Transformer to replace all the linear transformations or FC layers with PHM layers. The single-head self-attention module is rewritten as:
\begin{align*}
\Qb,\Kb,\Vb &= \Phi(\text{PHM}(\Xb)) \\
\Ab &= \text{softmax}(\frac{\Qb \Kb^\top}{\sqrt{d_k}})\Vb,
\end{align*}
where $d_k$ is the key dimension, $\Phi: \RR^{1  \times d} \rightarrow \RR^{3 \times \frac{d}{3}}$ is a three-way split on the last dimension, $\Xb$ is the input sequence, and $\Ab$ is the self-attentive representation. 
For multi-head attention, using PHM layers also enables weight sharing not only among the linear transformations of $\Qb,\Kb,\Vb$ but also among the linear transformation of multiple heads:
\begin{align*}
\Xb = \text{PHM}([\Hb_{1}; \ldots; \Hb_{N_{h}}]),    
\end{align*}
where $N_{h}$ is the number of heads and ($;$) is the column-wise concatenation. Finally, the position-wise feed-forward network is now defined as
\begin{align*}
\Yb = \text{PHM}(\text{ReLU}(\text{PHM}(\Xb))),    
\end{align*}
which transforms $\Xb$ with two PHM layers. 


\section{Experiments}

For context,
in the field of representation learning using hypercomplex multiplications, 
Quaternion convolutional neural networks~\citep{zhu2018quaternion}, 
Quaternion recurrent neural networks~\citep{parcollet2018quaternion2}, 
and Quaternion Transformers~\citep{tay2019lightweight}
have all
compared themselves with only real-valued counterparts. 
Therefore,
to be consistent with the rest of the literature,
we evaluate PHM-LSTMs and PHM-Transformers that are equipped with PHM layers,
and compare them with Quaternion LSTMs, Quaternion Transformers, real-valued LSTMs, or real-valued Transformers.
Both Quaternion LSTMs and Quaternion Transformers
replace linear transformations with Hamilton products of Quaternions.

To demonstrate the architectural flexibility and effectiveness,
we evaluate different settings of PHM-LSTMs 
and PHM-Transformers to show that
allowing for
flexible choices of the hyperparameter $n$ in the PHM layer
may lead to more effective performance. 
Details of the setup for the experiments are provided in Appendix~\ref{sec:Setup for Experiments}.

\subsection{Natural Language Inference}
The task of natural language inference is to determine the logical relationship between two text sequences~\citep{maccartney2009natural}. It is a fundamental task pertaining to language understanding. To this end, they serve as a suitable benchmark for evaluating recurrent models.

\begin{table*}
\centering
\caption{Experimental results of natural language inference (accuracy) on five different datasets. The PHM-LSTM reduces the parameters of the standard LSTM model and improves or partially matches performance on four out of five datasets.}
\label{nli}
\begin{tabular}{lcccccc}
\toprule
Model & \#Params &  MNLI & QNLI & SNLI & DNLI & SciTail \\
\midrule
LSTM & 721K & \textbf{71.82} / 71.89 & 84.44 & 84.18 & 85.16 & 74.36 \\
Quaternion LSTM & 180K (-75.0\%) &  71.57 / \textbf{72.19}  & \textbf{84.73} &  84.21 &  86.45 & 75.58  \\
\midrule
PHM-LSTM ($n=2$) &361K (-49.9\%) & \textbf{71.82} / 72.08 & 84.39 & 84.38 & 85.77 & 77.47 \\
PHM-LSTM ($n=5$) &146K (-79.7\%) & 71.80 / 71.77 & 83.87 & \textbf{84.58} & \textbf{86.47} & 74.64 \\
PHM-LSTM ($n=10$) &81K (-88.7\%) & 71.59 / 71.59 & 84.25 & 84.40 & 86.21 & \textbf{77.84} \\
\bottomrule
\end{tabular}
\end{table*}

We run experiments on five datasets: 
(i) MultiNLI~\citep{williams2017broad}, 
(ii) QNLI (Quora)~\citep{wang2017bilateral}, 
(iii) SNLI~\citep{bowman2015large}, 
(iv) Dialogue NLI~\citep{welleck2018dialogue}, 
and (v) SciTail (Science Entailment)~\citep{khot2018scitail}.
Table \ref{nli} reports the results on all these datasets. 
All in all, 
such results show that the PHM layer can not only
reduce the parameters but also improve performance
with flexible choices of $n$
(four out of five datasets show reasonable improvement or partially match).
The only exception
is on the QNLI dataset, 
where the performance drop is marginal ($< 1\%$).
This is still decent considering the parameter saving:
the parameterization cost of the PHM-LSTM
is in the order of $\mathcal{O}(1/n)$
of that of the standard LSTM,
where settings of $n=5$ and $n=10$ do not take values of power of 2.
As detailed in Appendix~\ref{sec:Setup for Experiments},
since we use the 300D GloVe~\citep{pennington2014glove} embeddings
to represent input tokens, 
we choose
multiples of 5 instead of 4 for ease of divisibility.
It is also noteworthy that
on the SNLI, Dialogue NLI, and SciTail datasets,
all the PHM-LSTM variants outperform the standard LSTM model.
We think that the 
element reusing properties of the Kronecker product operation,
in addition to learning to share such reused components
amongst recurrent gating functions,
may contribute to both effective and efficient representations.

\subsection{Machine Translation}
Machine translation is concerned with translating between source-target language pairs. To this end, sequence transduction models are central to this problem domain. In this experiment, the key goal is to compare PHM-Transformers against the standard and Quaternion Transformer models.

We run experiments on seven datasets: (i) IWSLT'15 English-Vietnamese (En-Vi), (ii) IWSLT'17 English-Indonesian (En-Id), (iii) IWSLT'14 German-English (De-En), (iv) IWSLT'14 Romanian-English (Ro-En), (v) WMT'18 English-Estonian (En-Et), (vi) Setimes English-Macedonian (En-Mk), and (vii) WMT'18 English-Romanian (En-Ro). 

Table~\ref{nmt} reports our results of the machine translation tasks. 
Overall, these empirical results
with different settings
demonstrate architectural flexibility and effectiveness
of the hypercomplex multiplication parameterization.
First and foremost, 
across six out of seven benchmarks, the PHM-Transformer at $n=4$ makes reasonable gains over the Quaternion Transformer,
signifying that parameterization of hypercomplex multiplications
by learning from data
can be more effective
than predefining Hamilton product rules mathematically.
Second, 
though increasing $n$ leads to more parameter savings,
we observe that increasing $n$ all the way to $16$ does not cause significant degradation in performance on datasets such as En-Vi.
Third, for most datasets, even with significant parameter savings, we find that the decrease in the BLEU score is mostly manageable ($\approx$ 1--3 BLEU points). However, we also note a rare occurrence where $n=16$ results in a significant decrease in the BLEU score, such as on the En-Id dataset. 
Fourth, on several datasets, the PHM-Transformer model improves the performance of the standard Transformer model.
For example, on datasets such as En-Vi and En-Et,
the PHM-Transformer model enjoys a performance boost of about $0.8$ BLEU point with $n=2$. Finally, by re-scaling with a factor of 2 (doubling the hidden size), we are able to improve the performance on three datasets: En-Vi, En-Id, and En-Mk. 


\begin{table}[t]
\small
\centering
\caption{Experimental results of machine translation (BLEU) on seven different datasets.
Symbol $\dagger$ represents re-scaling the parameters with a factor of 2
by doubling the hidden size.
The PHM-Transformer does not lose much performance despite enjoying parameter savings.
Re-scaling can lead to improvement in performance.}
\label{nmt}
\begin{tabular}{lcccccccc}
\toprule
Model & \#Params & En-Vi & En-Id  & De-En  & Ro-En  & En-Et & En-Mk & En-Ro \\
\midrule
Transformer (Tm) & 44M & 28.43 & 47.40 & \textbf{36.68} & \textbf{34.60} & 14.17 & 13.96 & \textbf{22.79}  \\
Quaternion Tm & 11M (-75.0\%) & 28.00 & 42.22 & 32.83 & 30.53 &13.10  & 13.67  & 18.50  \\
\midrule
PHM-Tm $n=2$ & 22M (-50.0\%) & 29.25 &	46.32 &	35.52 &	33.40 &	\textbf{14.98}	& 13.60 &	21.73  \\
PHM-Tm $n=4$ & 11M (-75.0\%) & 29.13 & 44.13 & 35.53 & 32.74 & 14.11 & 13.01 & 21.19  \\
PHM-Tm $n=8$ & 5.5M (-87.5\%) & 29.34 & 40.81 & 34.16 & 31.88 & 13.08 & 12.95 & 21.66  \\
PHM-Tm $n=16$ & 2.9M (-93.4\%) & 29.04 & 33.48 & 33.89 & 31.53 & 12.15 & 11.97 &	19.63  \\
\midrule
PHM-Tm$^\dagger$ $n=2$ & 44M & \textbf{29.54} & \textbf{49.05} & 34.32 & 33.88 & 14.05 & \textbf{14.41} & 22.18  \\
PHM-Tm$^\dagger$  $n=4$ & 22M (-50.0\%) & 29.17 & 46.24 & 34.86 & 33.80 & 14.43 & 13.78 & 21.91 \\
PHM-Tm$^\dagger$  $n=8$ & 11M (-75.0\%) & 29.47 & 43.49 & 34.71 &  32.59 & 13.75 &13.78 & 21.43 \\
\bottomrule
\end{tabular}
\end{table}

\begin{table}[t]
\centering
\caption{Training time (seconds per 100 steps) and inference time (seconds to decode test sets) with beam size of $4$ and length penalty of $0.6$ on the IWSLT'14 German-English dataset.}
\label{time}
\begin{tabular}{ccccc}
\toprule
Model &Transformer (Tm) &Quaternion Tm & PHM-Tm ($n=4$) & PHM-Tm ($n=8$)\\
\midrule
Training time & \textbf{7.61} &8.11 &7.92 & 7.70\\
Inference time & 336 &293 &299 &\textbf{282}\\
\bottomrule
\end{tabular}
\end{table}

Table \ref{time} reports the training and inference time for Transformer variants. We observe that the PHM-Transformer with $n=8$ has the fastest inference speed amongst all the variants, primarily due to a significant reduction of parameters.
All in all, the training speed is also approximately comparable.
This ascertains that the PHM layer 
does not increase much computational cost in practice.

\subsection{Text Style Transfer}
We continue to experiment with sequence transduction for text style transfer. The goal of this task is to convert text of a certain style to another style.
We use the Modern$\rightarrow$Shakespeare corpus\footnote{\url{https://github.com/tlatkowski/st}} in the experiments.
Table \ref{tst} reports the results on this text style transfer task. We observe that the best performance is achieved with the PHM-Transformer ($n=4$). Notably, all except the $n=16$ variant increases or matches the performance of the standard Transformer model. This ascertains  architectural flexibility and effectiveness of the proposed PHM layer. This not only enables parameter savings but also improves the performance of the Transformer.

\subsection{Subject Verb Agreement}
We conduct additional experiments on the subject-verb agreement task~\citep{linzen2016assessing}. The task predicts if the sentence, e.g., \textit{`The keys to the cabinet \_\_\_\_\_ .'} is followed by a plural or a singular. 
The used dataset can be found online~\citep{linzen2016assessing}.
Table~\ref{sva} reports the results on the subject-verb agreement task. Results are promising, demonstrating that all variants with PHM layers outperform the standard and Quaternion Transformer models. The best performance peaks at $n=8$, despite a parameter saving to up to $1/8$.

\section{Related Work}
While neural networks have been a well-established line of research, progress on hypercomplex representations for deep learning is still in its infancy and most works on this topic are  new~\citep{gaudet2017deep, parcollet2018quaternion2,parcollet2018quaternion,zhu2018quaternion, tay2019lightweight}.
The hypercomplex Hamilton product provides a greater extent of expressiveness, similar to the complex multiplication, albeit with a 4-fold increase in interactions between real and imaginary components. In the case of Quaternion representations, due to parameter savings in the Hamilton product, models also enjoy a 75\% reduction in the parameter size~\citep{parcollet2018quaternion2,tay2019lightweight}.
A striking caveat is that all Quaternions are fundamentally limited to 4D hypercomplex space, which restricts architectural flexibility.
The other options would be to scale to Octonion (8D) or Sedonion (16D) space,
given the predefined multiplication rules in such space.
To the best of our knowledge, there is no work that attempts to generalize arbitrary $n$D hypercomplex multiplications to allow for architectural flexibility,
where $n$ can be specified or tuned by users.


Our work can also be interpreted as a form of soft parameter sharing, albeit learned from data. 
Quaternion networks~\citep{zhu2018quaternion,parcollet2018quaternion,parcollet2019quaternion} are known to possess weight sharing properties via the Hamilton product operation and have demonstrated reasonable success despite having fewer parameters. To the best of our knowledge, there has been no work that attempts to parameterize the hypercomplex Hamilton product for neural networks, i.e., enabling end-to-end learning of real and imaginary component interactions from data.

\begin{table}[t]
\centering
\caption{Experimental results of text style transfer. The PHM-Transformer may reduce the parameters of the standard Transformer model and improve performance.}
\label{tst}
\begin{tabular}{ccc}
\toprule
Model & \#Params & BLEU \\
\midrule
Transformer (Tm) & 44M & 11.65 \\
\midrule
PHM-Tm ($n=2$) & 22M (-50.0\%) & 12.20  \\
PHM-Tm ($n=4$) & 11M (-75.0\%) & \textbf{12.42} \\
PHM-Tm ($n=8$) & 5.5M (-87.5\%) &11.66 \\
PHM-Tm ($n=16$) & 2.9M (-93.4\%) &10.76 \\
\bottomrule
\end{tabular}
\end{table}
\begin{table}[t]
\centering
\caption{Experimental results of subject verb agreement. The PHM-Transformer may reduce the parameters of the standard Transformer model and improve performance.}
\label{sva}
\begin{tabular}{ccc}
\toprule
Model &  \#Params & Acc \\
\midrule
Transformer (Tm) & 400K & 94.80 \\
Quaternion Tm & 100K & 94.70\\
\midrule
PHM-Tm ($n=2$) & 200K (-50.0\%) & 95.14 \\
PHM-Tm ($n=4$) & 101K  (-74.8\%) & 95.05 \\
PHM-Tm ($n=8$) & 56K  (-86.0\%) & \textbf{95.62} \\
\bottomrule
\end{tabular}
\end{table}

\section{Conclusion}
We proposed parameterized hypercomplex multiplication (PHM) layers that learn and generalize hypercomplex multiplications.
In practice,
the PHM layer has $1/n$ learnable parameters compared with the fully-connected layer counterpart, where $n$ can be flexibly specified by users.
PHM layers are applicable to dominant models such as LSTMs and Transformers. 
We evaluated these models equipped by PHM layers on comprehensive
tasks to show  
architectural flexibility and 
effectiveness of the hypercomplex multiplication parameterization.

\paragraph{Acknowledgements.} We thank the anonymous reviewers for the insightful comments on this paper. This work was partially supported by the Ministry of Education (MoE) of Singapore under the Academic Research Fund (AcRF) Tier 1 Grant RG135/18.

\bibliography{r2d2}
\bibliographystyle{iclr2021_conference}

\newpage

\appendix

\section{Reconstructing the Parameter Matrix}
\label{sec:Reconstructing the Parameter Matrix}

In the paper,
the parameter matrix $\Hb$ in \eqref{eq:PHM} is constructed by
a sum of $n$ Kronecker products.
In the following,
we will provide an alternative perspective and show how to
equivalently reconstruct $\Hb$
by reusing parameter matrices in real-valued matrix multiplications,
followed by more operations.

\subsection{Method}

The key idea is to operate on partitioned weight blocks and learn a dynamic diffusion of weights. There are two key parameter blocks $\Bb$ and $\Tb$ that are central to our approach. Intuitively, $\Bb \in \mathbb{R}^{n \times n \times n}$ controls the weight diffusion process and learns the soft interactions between $\Tb$ partitions. Here, $n$ is a user defined hyperparameter. 

Suppose that both $d$ and $k$ are divisible by $n \in \ZZ_{> 0}$. For $i = 1, \ldots, n$ and $j = 1, \ldots, \frac{d}{n}$, denote by each partitioned parameter block $\Tb_j \in \RR^{n \times \frac{k}{n}}$, and $\Bb_i \in \RR^{n \times n}$ is the weight diffusion matrix assigned to each partitioned parameter block via real-valued matrix multiplication $\Bb_i \Tb_j$.
The parameter $\Hb$ in \eqref{eq:PHM} is now constructed by column-wise concatenation ($;$):

\begin{align}
\label{eq:H_mat}
\Hb = [s(\Bb_1); s(\Bb_2); \ldots; s(\Bb_n)],
\end{align}

where each segment $s(\Bb_i)$ is also formed by column-wise concatenation:

\begin{align}
\label{eq:s_mat}
s(\Bb_i) = [\psi(\Bb_i \Tb_1); \psi(\Bb_i \Tb_2); \ldots; \psi(\Bb_i \Tb_{\frac{d}{n}})].
\end{align}

In \eqref{eq:s_mat}, function $\psi: \RR^{p \times q} \rightarrow \RR^{pq}$, where $\psi(\Xb)$ flattens the matrix $\Xb \in \RR^{p \times q}$ by concatenating each row of $\Xb$ then transposes the concatenated row vector into a column vector of dimension $pq$. 
It is easy to see that, $\psi(\Bb_i \Tb_j) \in \RR^k$, $s(\Bb_i) \in \RR^{k \times \frac{d}{n}}$, thus $\Hb \in \RR^{k \times d}$.

\begin{figure*}[t]
    \centering
  \includegraphics[width=0.95\linewidth]{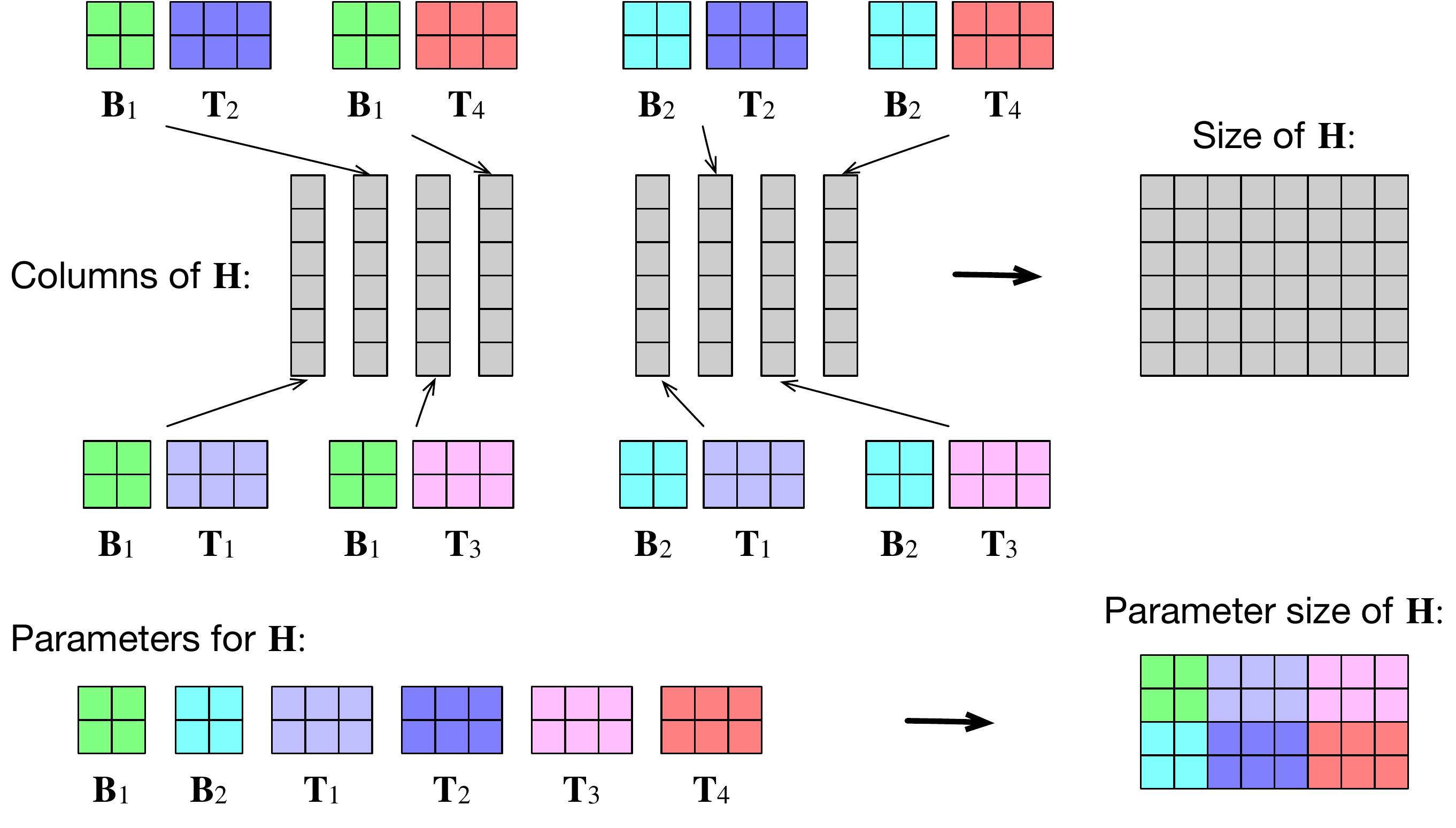}
  \vspace{-1em}
    \caption{Illustration of reconstructing $\Hb$ in \eqref{eq:PHM} by reusing parameter matrices $\Bb_i$ ($i=1,2$) and $\Tb_j$ ($j=1,\ldots,4$) in real-valued matrix multiplications, followed by more operations (here $n=2, k=6, d=8$). Best viewed in color.}
    \label{fig:paramsaving}
\end{figure*}

It is the partitioned parameter blocks $\Bb_i$ ($i = 1, \ldots, n$) and $\Tb_j$ ($j = 1, \ldots, \frac{d}{n}$) that determine the degree of freedom for $\Hb$, which is $kd/n + n^3$.
As illustrated in Figure~\ref{fig:paramsaving},
the reuse of parameter matrices 
$\Bb_1, \ldots, \Bb_n$ and $\Tb_1, \ldots, \Tb_{\frac{d}{n}}$
in real-valued matrix multiplications in \eqref{eq:s_mat}
may reduce the degree of freedom for $\Hb$.

\subsection{Subsuming Hypercomplex Multiplications}
\label{sec:hyper}

Similarly, we show how the PHM layer with the reconstructed $\Hb$ in \eqref{eq:H_mat}
also subsumes the hypercomplex multiplication.
Taking the Hamilton product of two Quaternions $Q$ and $P$ as an example,
it can be rewritten as
\begingroup\makeatletter\def\f@size{7}
\begin{align}
\label{eq:ASQ_mat}
\left(
\underbrace{
\begin{bmatrix}
    1 & 0 & 0 & 0 \\
    0 & 1 & 0 & 0 \\
    0 & 0 & 1 & 0 \\
    0 & 0 & 0 & 1 \\
\end{bmatrix}
}_{\Bb_1}
\underbrace{
\begin{bmatrix}
    Q_r \\
    Q_x \\
    Q_y \\
    Q_z \\
\end{bmatrix}
}_{\Tb_1};
\underbrace{
    \begin{bmatrix}
    0 & -1 & 0 & 0 \\
    1 & 0 & 0 & 0 \\
    0 & 0 & 0 & 1 \\
    0 & 0 & -1 & 0 \\
\end{bmatrix}
}_{\Bb_2}
\underbrace{
\begin{bmatrix}
    Q_r \\
    Q_x \\
    Q_y \\
    Q_z \\
\end{bmatrix}
}_{\Tb_1};
\underbrace{
    \begin{bmatrix}
    0 & 0 & -1 & 0 \\
    0 & 0 & 0 & -1 \\
    1 & 0 & 0 & 0 \\
    0 & 1 & 0 & 0 \\
\end{bmatrix}
}_{\Bb_3}
\underbrace{
\begin{bmatrix}
    Q_r \\
    Q_x \\
    Q_y \\
    Q_z \\
\end{bmatrix}
}_{\Tb_1};
\underbrace{
    \begin{bmatrix}
    0 & 0 & 0 & -1 \\
    0 & 0 & 1 & 0 \\
    0 & -1 & 0 & 0 \\
    1 & 0 & 0 & 0 \\
\end{bmatrix}
}_{\Bb_4}
\underbrace{
\begin{bmatrix}
    Q_r \\
    Q_x \\
    Q_y \\
    Q_z \\
\end{bmatrix}
}_{\Tb_1}
\right)
\begin{bmatrix}
    P_r \\
    P_x\\
    P_y\\
    P_z \\ 
\end{bmatrix},
\end{align}
\endgroup
where the 4 output elements are the real values for the Quaternion unit basis $[1, \ib, \jb, \kb]^\top$.
According to \eqref{eq:ASQ_mat}, when $n=4$,
the PHM layer with the reconstructed parameter matrix
can also be learned to exactly express the Hamilton product of Quaternions.
Likewise, hypercomplex multiplications of Octonions or Sedenions can also be learned by the PHM layer when $n$ is set to 8 or 16.

\subsection{Subsuming Real-Valued Matrix Multiplications}

Now we show how the PHM layer with the reconstructed $\Hb$ in \eqref{eq:H_mat}
also subsumes the matrix multiplication in real space.
Referring to \eqref{eq:PHM}, when $n=1$, $\Hb = b\Wb$, where the scalar $b$ is the single element of the $1\times 1$ matrix $\Bb_1$ and elements of $\Wb \in \RR^{k \times d}$ come from the concatenation of $\Tb_1, \ldots, \Tb_d \in \RR^{1\times k}$. Since learning $b$ and $\Wb$ separately is equivalent to learning their multiplication jointly, the scalar $b$ can be dropped,
which is learning the single weight matrix in an FC layer. Therefore, a PHM layer is degenerated to an FC layer when $n=1$.

\section{Setup for Experiments}
\label{sec:Setup for Experiments}

We describe the setup for the experiments as follows. 

\subsection{Natural Language Inference}

We implement 300D unidirectional encoders with shared parameters for both premises and hypotheses.
We take the concatenation of max and mean pooled representations as the input to a two-layer 300D multilayer perceptron for prediction.
Our model is trained with the the Adam optimizer with a learning rate of $0.0004$ and a batch size of $256$.
Word embeddings are initialized with GloVe~\citep{pennington2014glove} and are fixed.
No cross sentence attention~\citep{parikh2016decomposable} is used, mainly to observe the effectiveness of standalone encoders.
For PHM-LSTM, we use $n=\{2,5,10\}$.
Note that in this task, since word embeddings are 300D, we select multiples of $5$ instead of $4$ for ease of divisibility.

\subsection{Machine Translation} 

For the IWSLT'15 English-Vietnamese (En-Vi),
IWSLT'17 English-Indonesian (En-Id),
IWSLT'14 German-English (De-En),
and IWSLT'14 Romanian-English (Ro-En) datasets,
we run with 50K steps; 
while for 
WMT'18 English-Estonian (En-Et),
Setimes English-Macedonian (En-Mk),
and WMT'18 English-Romanian (En-Ro) datasets,
models are trained for 100K steps. 
For the En-Vi, En-Id, En-Et, En-Mk, and En-Ro datasets,
we specify that Transformers have 4 layers, $8$ heads, and a hidden size 512. 
For the De-En and Ro-En datasets, we specify that Transformers have 2 layers, $4$ heads, and a hidden size 256. 
We use beam size of $5$ and $\alpha=0.6$ (length penalty) for decoding.
For all PHM models, we benchmark several settings for the hyperparameter $n=\{2,4,8,16\}$. 


\subsection{Text Style Transfer}
For the used Modern$\rightarrow$Shakespeare corpus\footnote{\url{https://github.com/tlatkowski/st}} in the experiments, the key goal here is to convert modern writing into Shakespeare writing. This dataset comprises of $18,395$ parallel sentences for training, $1,218$ parallel sentences for evaluation (development set), and $1,462$ parallel sentences for testing. 
We still specify that Transformers have 4 layers, 8 heads, and a hidden size 512.
Similar to machine translation, we experiment with $n=\{2,4,8,16\}$. We train all the models for 10K steps.

\subsection{Subject Verb Agreement}


In contrast to the previous experimental settings, we use a smaller Transformer architecture with 10K training steps.
Specifically, Transformers here have 2 layers, 4 heads, and a hidden size 128.
Since the hidden size is smaller than those in the previous experimental settings,
we experiment with $n=\{2,4,8\}$.

\end{document}